\documentclass{article}

\usepackage[preprint]{neurips_2026}

\usepackage[utf8]{inputenc} 
\usepackage[T1]{fontenc}    
\usepackage{hyperref}       
\usepackage{url}            
\usepackage{booktabs}       
\usepackage{amsfonts}       
\usepackage{nicefrac}       
\usepackage{microtype}      
\usepackage{xcolor}         
\usepackage{enumitem}       
\usepackage{float}          
\usepackage{algorithm}      
\usepackage{algorithmic}    
\usepackage{amsmath,amssymb} 
\usepackage{tikz}           
\usetikzlibrary{positioning, arrows.meta, fit, backgrounds}

\title{Scaffold-Mediated Post-Training: Co-Evolving Model Parameters and Procedural Scaffold Graphs}

\author{%
  Fei Ding\thanks{Corresponding author: \texttt{dignfei@gmail.com}} \quad
  Yongkang Zhang \quad
  Runhao Liu \\
  Alibaba Group \\
  \And
  Yuhao Liao \quad
  Zijian Zeng \quad
  Huiming Yang \\
  Tsinghua University \\
}

\begin{document}

\maketitle

\begin{abstract}
Post-training of large language models optimizes only parameters, while inference-time procedural scaffolds are typically designed independently of parameter training. This disconnect makes it difficult to automatically acquire and internalize complex strategies. We propose \textbf{scaffold-mediated post-training}: procedural scaffolds are organized into an evolvable graph structure that co-evolves with model parameters through discovery, distillation, and dynamic recompilation. We instantiate this paradigm as Skill Training. On FeatureBench, automatically discovered skills improve the passed rate by 8.1pp, and after progressive distillation the model still achieves a 27.7\% passed rate without any external scaffold (distillation retention rate 85.2\%, defined as post-distillation / with-skill passed rate), significantly outperforming standard SFT on the same data.
\end{abstract}

\begin{figure}[H]
    \centering
    \begin{tikzpicture}[
        node distance=0.4cm,
        main/.style={rectangle, rounded corners=3pt, draw=#1!70, fill=#1!10,
            minimum width=2.0cm, minimum height=0.6cm,
            align=center, font=\scriptsize\bfseries, text=#1!65!black},
        main/.default=blue,
        sub/.style={font=\tiny},
        arr/.style={-{Stealth[length=2mm]}, semithick, #1},
        arr/.default=gray!70,
        lbl/.style={font=\tiny},
        center/.style={circle, draw=purple!50, fill=purple!5,
            minimum size=1.0cm, align=center, font=\tiny\bfseries, text=purple!70!black}
        ]

        \node[center] (core) {Co-\\evolution};

        \node[main=blue, above left=0.7cm and 1.4cm of core] (graph) {Scaffold Graph $G_t$};
        \node[main=blue, above right=0.7cm and 1.4cm of core] (data) {Data Generation};
        \node[main=teal, below right=0.7cm and 1.4cm of core] (model) {Model $M_{t+1}$};
        \node[main=red!60!black, below left=0.7cm and 1.4cm of core] (recomp) {Recompilation};

        \node[sub, text=blue!65!black, above=0.05cm of graph] (s1) {skill nodes+dep.\ edges};
        \node[sub, text=blue!65!black, above=0.05cm of data] (s2) {subgraph$\to$exec$\to$$V$-verify};
        \node[sub, text=teal!65!black, below=0.05cm of model] (s3) {bottom-up distill.+KL};
        \node[sub, text=red!65!black, below=0.05cm of recomp] (s4) {remove$\to$rewrite$\to$re-verify};

        \draw[arr=blue!70] (graph) -- (data)
        node[lbl, text=blue!65!black, midway, above]{$C$-select·$I$-assemble};
        \draw[arr=teal!70] (data) -- (model)
        node[lbl, text=teal!65!black, midway, right]{$\mathcal{L}+\beta D_{\text{KL}}$};
        \draw[arr=red!60] (model) -- (recomp)
        node[lbl, text=red!65!black, midway, below]{$\text{retain}_k\!\geq\!\tau$?};
        \draw[arr=purple!60] (recomp) -- (graph)
        node[lbl, text=purple!65!black, midway, left]{$G_{t+1}$};

        \node[sub, text=orange!70!black, left=0.2cm of graph] (seed) {Seed Data};
        \draw[arr, dashed] (seed) -- (graph);

    \end{tikzpicture}
    \caption{Co-evolution loop in scaffold-mediated post-training. $G_t$ assembles task-specific subgraphs and generates $V$-verified training data; the model internalizes skill behaviors via progressive distillation; skills meeting the retention threshold are removed and dependent scaffolds are recompiled into $G_{t+1}$.}
    \label{fig:overview}
\end{figure}
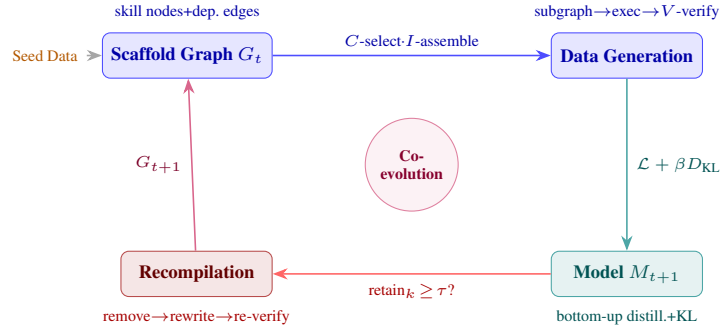

\section{Introduction}

\textbf{The parameter-learning bottleneck.} Post-training of large language models (LLMs) typically treats only model parameters as trainable. However, the procedural strategies required by complex tasks (how to decompose requirements, edit multiple files in topological order, and verify correctness after each step) are difficult to learn from scarce training examples alone~\citep{yang2025swesmithscalingdatasoftware}. Evaluations on FeatureBench~\citep{zhou2026featurebench} show that Claude Opus 4.5 achieves a 76.8\% solve rate on SWE-bench~\citep{jimenez2024swebench} yet only a 45.5\% passed rate on FeatureBench, revealing deficiencies in procedural strategies. FeatureBench focuses on end-to-end feature development (involving multi-file coordination, architectural decisions, and multi-step verification workflows), placing higher demands on procedural strategies.

\textbf{The limits of scaffold engineering.} Inference-time scaffolds such as chain-of-thought~\citep{NEURIPS2022_9d560961}, ReAct~\citep{yao2023react}, and agentic workflows can substantially improve performance on complex tasks, but they are typically hand-designed and bolted on at inference time; as scaffolds accumulate, inference becomes increasingly complex and slower.

\textbf{Scaffold-mediated post-training.} This paper proposes a new post-training paradigm that bridges parameter learning and scaffold engineering: \textbf{procedural scaffolds themselves become evolvable objects that co-evolve with model parameters}. The model first solves tasks with the aid of an external procedural scaffold graph, which generates high-quality supervised data. The model then absorbs the scaffold's behavior through distillation. Once a scaffold has been internalized into parameters, the external graph undergoes \textbf{dynamic recompilation}: the internalized skill is removed from the graph, and higher-level scaffolds that depend on it are automatically rewritten, closing the co-evolution loop between parameter learning and scaffold learning.

We instantiate this paradigm as \textbf{Skill Training}. A skill is a node in the procedural scaffold graph, comprising metadata ($\textit{meta}$), an activation condition ($C$), an execution procedure ($P$), a verification checkpoint ($V$), a compositional interface ($I$), and anti-patterns ($E$). The framework operates in three stages (Figure~\ref{fig:overview}):

\begin{enumerate}[leftmargin=*]
    \item \textbf{Skill Discovery}: Structured skills are automatically discovered from seed data to construct an initial procedural scaffold graph.
    \item \textbf{Scaffold-Graph-Augmented Data Generation}: For each task, a scaffold subgraph is assembled to guide the model through the task and generate verified training data.
    \item \textbf{Progressive Distillation with Dynamic Recompilation}: Starting from the most frequently invoked bottom-level skills, progressive distillation internalizes skill behavior into model parameters. Internalized skills are removed from the scaffold graph, and higher-level scaffolds that depend on them are automatically rewritten and re-verified.
\end{enumerate}

The main contributions of this paper are:

\begin{itemize}[leftmargin=*]
    \item \textbf{A new paradigm}: We propose scaffold-mediated post-training, in which model parameters and an external procedural scaffold graph co-evolve through discovery, assembly, distillation, and dynamic recompilation.
    \item \textbf{A new representation}: We formalize skills as nodes in a procedural scaffold graph $G=(S,D)$, each containing the six-tuple $(\textit{meta}, C, P, V, I, E)$. Explicit skill calls in $P$ induce dependency edges in the graph, $I$ defines typed labels for inter-node data flow, and $V$ provides quality-filtering signals for distillation.
    \item \textbf{A new mechanism}: We propose dynamic recompilation: once a skill has been distilled and internalized into parameters, the scaffold graph automatically rewrites higher-level scaffolds that depend on it, enabling co-evolution between parameters and scaffolds.
    \item \textbf{Empirical validation}: On FeatureBench, automatically discovered skills raise the passed rate from 24.4\% to 32.5\%; after progressive distillation the rate reaches 31.5\% (22 skills internalized); pure internalization without external scaffolds still achieves 27.7\% (retention rate 85.2\%), all significantly outperforming standard SFT on the same data (25.5\%). Ablations show that $V$ and $P$ contribute 6.9pp and 4.2pp to the retention rate, respectively.
\end{itemize}

\noindent Limitations remain: this work validates the framework only in code generation and uses self-distillation from a single model without exploring teacher diversity. The framework itself is domain-agnostic, and extending it to other domains with executable verifiers, such as mathematical reasoning, is an important direction for future work (Section~\ref{sec:limitations}).

\section{Related work}

\subsection{Parameter learning}

SFT and RLHF compress capabilities directly into model parameters. STaR~\citep{NEURIPS2022_639a9a17} and Quiet-STaR~\citep{zelikman2024quietstar} achieve self-improvement by bootstrapping reasoning chains. Context Distillation~\citep{snell2022learningdistillingcontext} distills rich contexts into weights. These methods can update parameters but lack explicit, composable, and verifiable external procedural structures; their distillation targets are flat text or reasoning chains that lack explicit composable interfaces for programmatic cross-task reuse.

\subsection{Scaffold learning}

APE~\citep{zhou2023large} and OPRO~\citep{ICLR2024_3339f19c} automatically optimize prompts; Promptbreeder~\citep{pmlr-v235-fernando24a} optimizes prompts via evolutionary algorithms, yet all produce flat text. Voyager~\citep{wang2024voyager} builds a composable skill library; ExpeL~\citep{Zhao_Huang_Xu_Lin_Liu_Huang_2024} extracts natural-language rules; SkillRL~\citep{xia2026skillrlevolvingagentsrecursive} jointly trains skills with RL; Skill-It!~\citep{NEURIPS2023_70b8505a} studies data dependencies among skills. DSPy~\citep{ICLR2024_f1cf02ce} treats prompts as composable modules; SAGE~\citep{wang2026reinforcementlearningselfimprovingagent} accumulates a skill library within an RL framework. These methods construct external scaffolds but all \textbf{lack bidirectional co-evolution between parameters and the scaffold graph}.

\subsection{Bridging the two: scaffold-mediated post-training}

This paper connects parameter learning and scaffold learning: procedural scaffolds are first discovered and used to guide the model in solving tasks and generating training data (the strength of scaffold learning), then distilled and internalized into parameters (the strength of parameter learning), after which the scaffold graph undergoes dynamic recompilation. This ``learn $\rightarrow$ internalize $\rightarrow$ recompile'' loop has not been realized in prior work.

\section{Method: scaffold-mediated post-training}

\textbf{Engineering motivation.} In engineering practice, we built a system of 136+ manually authored skills covering Test-Driven Development (TDD), debugging, and code review workflows. These skills closely match the formal definition in structure (Appendix~\ref{app:skills}), but manual design is prohibitively expensive, directly motivating the automated training research below.

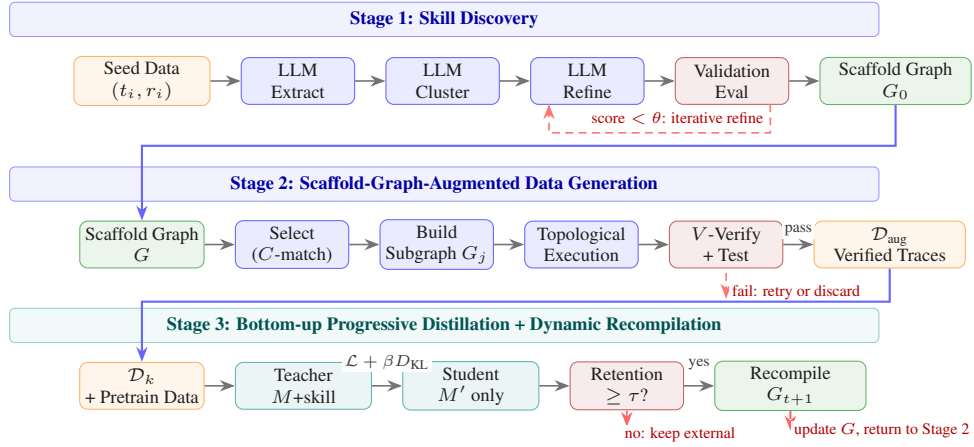
\begin{figure*}[t]
    \centering
    \begin{tikzpicture}[
        node distance=0.42cm and 0.4cm,
        box/.style={rectangle, rounded corners=3pt, draw=#1!60, fill=#1!8,
            minimum height=0.62cm, align=center, font=\scriptsize,
            text=black!88, inner sep=2pt},
        box/.default=blue,
        lbl/.style={font=\tiny, text=black!78, fill=white, fill opacity=0.92,
            text opacity=1, inner sep=1pt},
        note/.style={font=\tiny, text=red!70!black, fill=white, fill opacity=0.95,
            text opacity=1, inner sep=1pt, align=center},
        arr/.style={-{Stealth[length=2mm]}, semithick, draw=black!55},
        darr/.style={-{Stealth[length=2mm]}, densely dashed, semithick, #1},
        darr/.default=red!55,
        phase/.style={rectangle, rounded corners=2pt, draw=#1!40, fill=#1!5,
            minimum width=11.5cm, minimum height=0.32cm,
            font=\scriptsize\bfseries, text=#1!65!black},
        phase/.default=blue
        ]

        \node[phase=blue] (t1) {Stage 1: Skill Discovery};
        \node[box=orange, below=0.25cm of t1, xshift=-4cm, minimum width=1.8cm] (seed) {Seed Data\\$(t_i,r_i)$};
        \node[box=blue, right=0.4cm of seed, minimum width=1.5cm] (ext) {LLM\\Extract};
        \node[box=blue, right=0.4cm of ext, minimum width=1.5cm] (clu) {LLM\\Cluster};
        \node[box=blue, right=0.4cm of clu, minimum width=1.5cm] (ref) {LLM\\Refine};
        \node[box=red!60!black, right=0.4cm of ref, minimum width=1.5cm] (eval1) {Validation\\Eval};
        \node[box=green!50!black, right=0.4cm of eval1, minimum width=2.0cm] (g0) {Scaffold Graph\\$G_0$};
        \draw[arr] (seed) -- (ext);
        \draw[arr] (ext) -- (clu);
        \draw[arr] (clu) -- (ref);
        \draw[arr] (ref) -- (eval1);
        \draw[arr] (eval1) -- (g0);
        \draw[darr] ([xshift=-0.25cm,yshift=0.01cm]eval1.south east) -- ++(0,-0.38) coordinate (fb1)
        -- node[note, pos=0.42, above, yshift=2pt]{\tiny $\text{score}<\theta$: iterative refine}
        ([xshift=0.25cm]ref.south west |- fb1) -- ([xshift=0.25cm,yshift=0.01cm]ref.south west);

        \node[phase=blue, below=0.80cm of seed, xshift=4cm] (t2) {Stage 2: Scaffold-Graph-Augmented Data Generation};
        \node[box=green!50!black, below=0.25cm of t2, xshift=-4cm, minimum width=1.5cm] (g2) {Scaffold Graph\\$G$};
        \node[box=blue, right=0.4cm of g2, minimum width=1.5cm] (sel) {Select\\($C$-match)};
        \node[box=blue, right=0.4cm of sel, minimum width=1.5cm] (sub) {Build\\Subgraph $G_j$};
        \node[box=blue, right=0.4cm of sub, minimum width=1.5cm] (exec) {Topological\\Execution};
        \node[box=red!60!black, right=0.4cm of exec, minimum width=1.5cm] (vfy) {$V$-Verify\\+ Test};
        \node[box=orange, right=0.4cm of vfy, minimum width=2.0cm] (daug) {$\mathcal{D}_{\text{aug}}$\\Verified Traces};
        \draw[arr] (g2) -- (sel);
        \draw[arr] (sel) -- (sub);
        \draw[arr] (sub) -- (exec);
        \draw[arr] (exec) -- (vfy);
        \draw[arr] (vfy) -- (daug) node[lbl, midway, above, yshift=2pt]{\tiny pass};
        \draw[darr] ([yshift=-0.06cm]vfy.south) -- ++(0,-0.34)
        node[note, near end, right=1pt]{\tiny fail: retry or discard};

        \node[phase=teal, below=0.52cm of g2, xshift=4cm] (t3) {Stage 3: Bottom-up Progressive Distillation + Dynamic Recompilation};
        \node[box=orange, below=0.25cm of t3, xshift=-4cm, minimum width=1.5cm] (d3) {$\mathcal{D}_k$\\+ Pretrain Data};
        \node[box=teal, right=0.4cm of d3, minimum width=1.8cm] (teach) {Teacher\\$M$+skill};
        \node[box=teal, right=0.4cm of teach, minimum width=1.8cm] (stu) {Student\\$M'$ only};
        \node[box=red!60!black, right=0.4cm of stu, minimum width=1.5cm] (ret) {Retention\\$\geq\tau$?};
        \node[box=green!50!black, right=0.4cm of ret, minimum width=2.0cm] (recomp) {Recompile\\$G_{t+1}$};
        \draw[arr] (d3) -- (teach);
        \draw[arr] (teach) -- (stu) node[lbl, midway, above, yshift=5pt]{\tiny $\mathcal{L}+\beta D_{\text{KL}}$};
        \draw[arr] (stu) -- (ret);
        \draw[arr] (ret) -- (recomp) node[lbl, midway, above, yshift=5pt]{\tiny yes};
        \draw[darr=red!60] (ret.south) -- ++(0,-0.22)
        node[note, below right=-1pt and -2pt]{\tiny no: keep external};

        \draw[arr, blue!60, line width=0.8pt] (g0.south) -- ++(0,-0.54) -| (g2.north);
        \draw[arr, blue!60, line width=0.8pt] (daug.south) -- ++(0,-0.44) -| (d3.north);
        \draw[darr=red!60, line width=0.8pt] (recomp.south) -- ++(0,-0.26)
        node[note, right]{\tiny update $G$, return to Stage 2};

    \end{tikzpicture}
    \caption{Detailed three-stage data flow. \textbf{Stage 1}: iteratively discover skills from seed data and build the scaffold graph $G_0$. \textbf{Stage 2}: for each task, select nodes by $C$-matching, build subgraph $G_j$, execute in topological order, and $V$-verify; only passing traces are retained. \textbf{Stage 3}: bottom-up progressive distillation (loss includes KL constraint, mixed with pretrain data); skills whose retention rate meets the threshold are removed from the graph and trigger dependency recompilation. Blue arrows denote cross-stage data flow; red dashed lines denote feedback/fallback paths.}
    \label{fig:method}
\end{figure*}

This section first defines the procedural scaffold graph (\S3.1), then describes three stages: skill discovery (\S3.2) $\rightarrow$ scaffold-graph-augmented data generation (\S3.3) $\rightarrow$ progressive distillation and dynamic recompilation (\S3.4--3.5). The first two stages constitute \textbf{scaffold learning}; the third constitutes \textbf{parameter learning and scaffold-graph co-evolution} (Figure~\ref{fig:method}).

\subsection{Procedural scaffold graph}

\textbf{Definition 1} (Skill). A skill $s$ is a node in the procedural scaffold graph, comprising the following core components:
\[s = (\textit{meta}, C, P, V, I, E)\]
where:
\begin{itemize}[leftmargin=*]
    \item $\textit{meta}$ (metadata): name, description, category tags, available tool list, and version number.
    \item $C$ (activation condition): task characteristics and triggering rules that determine when the skill is activated (the LLM automatically evaluates the match against the task description and executes upon match).
    \item $P$ (execution procedure): structured workflow steps, categorized as either \textbf{rigid} (strict step ordering with hard constraints at each step) or \textbf{flexible} (advisory patterns that permit discretionary step selection).
    \item $V$ (verification checkpoint): checkable conditions that must be satisfied after skill execution. Following the \textbf{evidence-driven} principle, $V$ requires executable evidence rather than model assertions. Verification may comprise multiple layers (build $\rightarrow$ type-check $\rightarrow$ test $\rightarrow$ security $\rightarrow$ change audit); in rigid skills these are executed sequentially with fail-fast semantics (Appendix~\ref{app:verification}).
    \item $I = (I.\text{in}, I.\text{out})$ (compositional interface): standardized input/output type label sets, enabling multiple skills to be chained according to dependency relations.
    \item $E$ (anti-patterns): common failure modes that help the model avoid known pitfalls.
\end{itemize}

In practice, each skill is implemented as a Markdown file with YAML frontmatter; the design derives from iterative experience with over 136 manually authored skills.

\textbf{Definition 2} (Procedural Scaffold Graph). A scaffold graph $G = (S, D)$, where $S$ is the set of skill nodes and $D$ is the set of dependency edges induced by explicit skill calls in $P$: if $P_b$ explicitly invokes $s_a$, an edge from $s_a$ to $s_b$ is added. $I$ provides typed labels (e.g., \texttt{file\_list}, \texttt{test\_result}) for the data flowing along these edges at runtime. Acyclicity is verified via depth-first search (DFS) after graph construction: if a node reappears in the path stack, the cycle is identified and its chain reported, the offending last edge is discarded, and the check is retried. The graph of 30 skills in our experiments is always a DAG. Composite skills correspond to executing a scaffold subgraph: nodes are scheduled in topological order. $V$ provides verification feedback after each node executes.

Multi-round discovery produces hierarchical skill structures: higher-level skills discovered in later rounds explicitly invoke lower-level skills in their $P$, from which dependency edges naturally arise; Algorithm~\ref{alg:discovery} formalizes these call relationships via interface matching ($I_a.\text{out} \cap I_b.\text{in} \neq \emptyset$).

\textbf{Key distinction from prompts.} A conventional prompt is flat text; skills possess three structural properties: (1)~\textbf{Verifiability}---$V$ enables automatic evaluation of execution outcomes, providing a filtering signal for distillation; (2)~\textbf{Composability}---$I$ defines standardized interfaces that allow multiple skills to be chained; (3)~\textbf{Conditional activation}---$C$ automates skill selection. Ablation experiments (Section~\ref{sec:ablation}) quantify the contributions of $V$ and $I$.

In practice, skills are classified as either \textbf{rigid} (enforcing step ordering and verification checkpoints) or \textbf{flexible} (advisory patterns permitting flexible selection). Ablation experiments (Section~\ref{sec:ablation}) progressively remove $V$ and $I$ to verify the contribution of skill's core structural features ($E$ removal causes only 0.8pp drop, not significant). Full ablation configurations and skill examples are provided in Appendix~\ref{app:skills}.

\subsection{Stage 1: skill discovery}

\textbf{Input}: A high-quality seed dataset $\mathcal{D}_{\text{seed}} = \{(t_i, r_i)\}_{i=1}^{N}$, where $t_i$ is a task description and $r_i$ is a successful solution. $N$ is typically 1000--3000.

Algorithm~\ref{alg:discovery} (Appendix~\ref{app:algorithms}) provides the complete procedure. Here $\text{LLM.Extract}$, $\text{LLM.Cluster}$, and $\text{LLM.Refine}$ are all LLM calls (prompt templates are given in Appendix~\ref{app:prompts}). $\text{Evaluate}$ compares the passed rate on a validation set before and after skill injection. The threshold $\theta$ (default 0.05, i.e., a skill must yield $\geq$5\,pp improvement on the tasks that invoke it) controls the minimum effectiveness requirement; $\epsilon$ (default 0.01) controls the iteration convergence criterion. Sensitivity analysis is provided in Appendix~\ref{app:sensitivity}.

\textbf{Relationship to human-designed skills}. The skills produced by automatic discovery are structurally highly similar to manually designed ones (Appendix~\ref{app:skills} provides a component-by-component comparison), yet the process is fully automated, demonstrating that LLMs can induce structured strategies from successful exemplars.

\subsection{Stage 2: scaffold-graph-augmented data generation}

\textbf{Objective}: Leverage the scaffold graph $G_0$ to guide the model in solving additional tasks and generate high-quality training data. $\mathcal{T}_{\text{new}}$ comprises 2000 task instances generated by the FeatureBench data pipeline on additional repositories; success is determined by execution-based verification (running the test suite). Algorithm~\ref{alg:datagen} (Appendix~\ref{app:algorithms}) provides the procedure.

\textbf{Subgraph construction and execution.} The LLM selects candidate nodes $\mathcal{S}_j$ by $C$-matching, extracts subgraph $G_j$ from $G_0$ in topological order (depth limited to 3 layers). During execution, nodes are dispatched sequentially with data passed via $I$; each node runs $V$-verification after execution (one retry permitted; second failure discards the trajectory). A final task-level test suite is run; only fully verified trajectories are retained as training data. Formal definitions are given in Appendix~\ref{app:composition}.

\subsection{Stage 3: progressive distillation and dynamic recompilation}

\textbf{Objective}: Internalize skill behaviors into model parameters and dynamically recompile the scaffold graph after internalization. Algorithm~\ref{alg:distill} (Appendix~\ref{app:algorithms}) provides the complete procedure.

\textbf{Distillation objective}. We adopt a self-distillation approach in which the teacher and student are different configurations of the same model: the teacher receives $[\text{skill prompt} \oplus t_j]$ and produces $r_j$ (the full execution trajectory including intermediate reasoning and action steps); the student receives only $[t_j]$ and is trained to match $r_j$. The training loss is $\mathcal{L} + \beta \cdot D_{\text{KL}}(\pi_\theta \| \pi_{\text{ref}})$, where $\mathcal{L}$ is the cross-entropy loss and the KL divergence constraint prevents the parameters from deviating too far from the reference model.

\textbf{Distillation order}. Progressive distillation proceeds bottom-up along the topological order of the scaffold graph $G$: root nodes with no dependencies (in-degree 0) are distilled first; within the same level, nodes are sorted in descending order of dependency count (out-degree), prioritizing internalization of foundational capabilities with the largest impact. This ordering is uniquely determined by the graph structure and requires no manual weight tuning.

\textbf{Implementation details}. We use full-parameter fine-tuning (Appendix~\ref{app:cost}). The retention rate threshold is $\tau=0.85$, and the recompilation fallback threshold is $\delta=0.02$. The reference policy $\pi_{\text{ref}}$ in the KL constraint is statically anchored to the original base model $M_0$ and is not updated across distillation rounds. Algorithm~\ref{alg:distill} iterates per skill for clarity; in implementation, skills within the same topological level are merged into a single training round to avoid redundant gradient steps.

\textbf{Catastrophic forgetting mitigation.} Five mechanisms jointly mitigate catastrophic forgetting: (1)~\textbf{self-distillation}---teacher and student share the same base model, inherently small KL divergence reduces distribution shift; (2)~\textbf{bottom-up implicit replay}---lower-level skill behaviors are repeatedly replayed in higher-level training data; (3)~\textbf{pretraining data mixing}---50\% pretrain data mixed per round; (4)~\textbf{KL constraint} ($\beta=0.01$); (5)~\textbf{per-round retention check}---skills falling below $\tau$ are not internalized and are retained as external scaffolds. Detailed analysis is provided in Appendix~\ref{app:forgetting}.

\subsection{Dynamic recompilation}
\label{sec:recompilation}

Dynamic recompilation enables co-evolution of parameters and the scaffold graph. After skill $s_k$ is internalized, its node cannot simply be removed---higher-level skills may depend on it. Recompilation consists of six steps: (1)~\textbf{Node Removal}: remove $s_k$ from $G$; (2)~\textbf{Procedure Rewriting}: an LLM replaces explicit calls to $s_k$ in the higher-level scaffold $s_h$'s procedure $P_h$ with implicit assumptions about the model's internalized capability; (3)~\textbf{Bridge Regeneration}: regenerate inter-node data-passing instructions; (4)~\textbf{Interface Update}: update $I_h$; (5)~\textbf{Re-verification}: recompiled scaffolds must pass $V$ verification on a validation set (60s timeout per node); (6)~\textbf{Fallback}: if performance drops by $>\delta$, retain $s_k$ as an external scaffold. Recompilation depth is limited to 1 hop; each node is processed independently.

Formally: $M_{t+1} = \text{Distill}(M_t, \mathcal{D}(s_t))$; $G_{t+1} = \text{Recompile}(G_t, s_t, M_{t+1})$.

\section{Experimental setup}

Our experiments aim to answer five research questions: \textbf{RQ1}: Can a scaffold graph be automatically learned from seed trajectories? \textbf{RQ2}: Can the scaffold graph improve task solving and data generation? \textbf{RQ3}: Can scaffold behaviors be internalized into model parameters? \textbf{RQ4}: Is the graph structure ($V$ and $I$) necessary? \textbf{RQ5}: Is dynamic recompilation effective?

\subsection{Benchmarks and models}
\label{sec:benchmarks}

\textbf{Data and Evaluation.} We use the open-source data pipeline from FeatureBench~\citep{zhou2026featurebench} to generate training data. Evaluation strictly follows the FeatureBench paper setup: we use its standard Full set (200 tasks, 24 open-source repositories, execution-based verification), with OpenHands~\citep{wang2025openhands} as the base agent framework (providing tool interaction and environment management), and skills injected as an additional procedural knowledge layer on top. We report Passed Rate (the average proportion of fail-to-pass tests passed per task). Training data is generated using the data pipeline on additional repositories with no overlap with the evaluation set.

\textbf{Models.} \textbf{Qwen3-Coder-480B-A35B-Instruct}~\citep{yang2025qwen3technicalreport} (MoE, 35B activated parameters, full-parameter fine-tuning) and \textbf{DeepSeek-V3.2}~\citep{deepseekai2025deepseekv32} (MoE, 37B activated parameters, to verify cross-model generality).

\subsection{Experimental configurations}

\begin{table}[t]
    \centering
    \caption{Experimental configurations. Direct Fine-tune trains on the same $\mathcal{D}_{\text{aug}}$ via standard SFT in a single pass, isolating the contribution of structured training pathways.}
    \label{tab:configs}
    \small
    \begin{tabular}{lp{8.5cm}}
        \toprule
        \textbf{Configuration} & \textbf{Description} \\
        \midrule
        Baseline & Vanilla model executes tasks directly \\
        + Human Skills & Manually designed skill set \\
        + Auto Skills & Model auto-discovered skills \\
        + OPRO Prompt + Distill & OPRO-optimized prompt distilled; no prompt injected at inference \\
        + Distilled (All-at-once) & All-at-once distillation ($\tau$=0.85, 12 internalized + 18 external) \\
        + Distilled (Progressive) & Progressive distillation ($\tau$=0.85, 22 internalized + 8 external) \\
        + Pure Internalization & Progressive distillation with all external skills removed (0 external, parameters only) \\
        + Direct Fine-tune & Standard SFT on the same $\mathcal{D}_{\text{aug}}$ (with identical pretrain data mixing, no skill decomposition, no progressive curriculum) \\
        \bottomrule
    \end{tabular}
\end{table}

\subsection{Seed data construction}

We use the FeatureBench data pipeline to generate training instances on additional repositories, selecting 2000 seed tasks and 200 validation tasks (for skill quality evaluation, independent of the FeatureBench evaluation set). Selection criteria: involving multi-file modifications, containing strategic steps, and covering diverse development patterns. An additional 2000 tasks are sampled as $\mathcal{T}_{\text{new}}$ for the data generation phase. Seed data details are provided in Appendix~\ref{app:seeds}.

\subsection{Evaluation metrics}

\begin{itemize}[leftmargin=*]
    \item \textbf{Passed Rate}: the average proportion of fail-to-pass tests passed per task. We adopt Passed Rate as the primary metric because skills provide incremental help on procedural strategies (e.g., correctly executing 3 out of 5 steps), and Passed Rate captures such partial improvements. The strict Resolved Rate (all tests must pass) is reported in Appendix~\ref{app:resolved}; trends are consistent.
    \item \textbf{Distillation retention rate}: Passed Rate of the distilled model without skill / Passed Rate of the pre-distillation base model with skill (i.e., the Auto Skills passed rate of 32.5\%), measuring how much of the skill-augmented capability is retained after distillation (evaluation tasks are highly relevant to the skills).
    \item \textbf{Skill transfer rate}: the proportion of skills that yield a positive improvement on tasks outside the seed set.
\end{itemize}

Following the FeatureBench evaluation protocol, all 200 evaluation tasks are used for assessment (model training data comes entirely from independent seed and generated sets with no overlap with these 200 evaluation tasks). Each configuration is run with 5 random seeds; we report mean $\pm$ standard deviation.

\section{Experimental results}

\subsection{Main results (RQ1--RQ3)}

Table~\ref{tab:main} presents the passed rate of each configuration on FeatureBench.

\begin{table}[t]
    \centering
    \caption{Passed rate (\%) on FeatureBench Full (200 tasks, OpenHands framework, 5 random seeds).}
    \label{tab:main}
    \begin{tabular}{lcc}
        \toprule
        \textbf{Configuration} & \textbf{Qwen3-Coder} & \textbf{DeepSeek-V3.2} \\
        \midrule
        Baseline & $24.4 \pm 0.8$ & $26.1 \pm 0.7$ \\
        + Human Skills & $35.0 \pm 1.0$ & $36.8 \pm 1.1$ \\
        + Auto Skills & $32.5 \pm 1.1$ & $34.3 \pm 0.8$ \\
        + OPRO Prompt + Distill & $26.0 \pm 0.8$ & $27.8 \pm 0.6$ \\
        + Distilled (All-at-once) & $29.0 \pm 0.9$ & $30.4 \pm 1.0$ \\
        + Distilled (Progressive) & $31.5 \pm 1.0$ & $33.3 \pm 0.7$ \\
        + Pure Internalization & $27.7 \pm 1.0$ & $29.5 \pm 0.8$ \\
        + Direct Fine-tune & $25.5 \pm 0.9$ & $27.2 \pm 0.8$ \\
        \bottomrule
    \end{tabular}
\end{table}

\textbf{Key observations}.

(1)~\textbf{Effectiveness of auto-discovery}: Auto Skills (32.5\%) approach Human Skills (35.0\%) with a gap of only 2.5pp, indicating that the model can automatically discover skills from 2{,}000 seed cases at near-human quality (achieving 93\% of the human-designed passed rate).

(2)~\textbf{Value of scaffold-augmented inference}: Auto Skills (32.5\%) significantly outperform Baseline (24.4\%) by 8.1pp, demonstrating that skill-augmented inference yields a substantial absolute improvement.

(3)~\textbf{Value of distillation internalization}: Progressive Distilled (31.5\%, with 8 external skills) significantly outperforms Direct Fine-tune (25.5\%) by 6.0pp. Even after completely removing external skills, the purely internalized model still reaches 27.7\% (retention rate 85.2\%), outperforming Direct Fine-tune by 2.2pp. Note that both use \textbf{the same} $\mathcal{D}_{\text{aug}}$ (all $V$-verified); the gap arises purely from structured training pathways: per-skill progressive distillation vs.\ one-pass standard SFT.

(4)~Trends are consistent across both models.

\subsection{Skill quality analysis}

The automatic discovery process extracted approximately 120 candidate strategies from 2{,}000 seed cases, which were clustered and merged into 45 skill candidates; 30 of these passed validation-set filtering ($\theta=0.05$), converging in 5 iterations.

Auto Skills cover 78\% of test tasks (Human Skills cover 85\%), with a false-trigger rate of 12\% (Human Skills: 8\%). On out-of-seed tasks, 27 out of 30 skills yield a positive improvement (skill transfer rate 90\%), indicating good cross-task transferability. A fine-grained analysis stratified by task complexity is provided in Appendix~\ref{app:finegrained}.

\subsection{Graph structure ablation (RQ4)}
\label{sec:ablation}

\begin{table}[t]
    \centering
    \caption{Skill component ablation (Qwen3-Coder). Pure internal.\ rate = infer.\ passed rate $\times$ retention rate. All rows use the progressive distillation pipeline; Flat Prompt's low retention reflects pipeline incompatibility with unstructured prompts. OPRO+Distill in Table~\ref{tab:main} uses standard context distillation.}
    \label{tab:ablation}
    \small
    \resizebox{\columnwidth}{!}{%
    \begin{tabular}{lcccc}
        \toprule
        \textbf{Configuration} & \textbf{Infer.\ Passed Rate} & \textbf{Pure Internal.\ Rate} & \textbf{Retention Rate} & \textbf{Composable} \\
        \midrule
        Full Skill (complete) & 32.5\% & 27.7\% & 85.2\% & Yes \\
        $-V$ (no verification) & 30.0\% & 23.5\% & 78.3\% & Yes \\
        $-P$ (no explicit calls) & 31.0\% & 25.1\% & 81.0\% & No \\
        $-V-P$ (cond.\ prompt) & 28.5\% & 20.9\% & 73.3\% & No \\
        Flat Prompt (OPRO) & 27.0\% & 19.0\% & 70.4\% & No \\
        \bottomrule
    \end{tabular}%
    }
\end{table}

$V$'s dual role: removing $V$ drops inference-time passed rate by 2.5pp and retention rate by 6.9pp; $V$ provides execution feedback during inference and data filtering during distillation. $P$'s indirect value: removing explicit calls in $P$ drops passed rate by 1.5pp; the dependency graph induced by $P$ enables skill composition and makes distillation targets more modular ($-P$ leaves no dependency edges; the model autonomously decides which skill to invoke, and all skills are merged into a single round of joint distillation). Full degradation to conditional prompt ($-V-P$) differs from Flat Prompt by only 1.5pp in inference-time passed rate (28.5\% vs 27.0\%), confirming that $V$ and $P$ are the core distinguishing features.

\subsection{Dynamic recompilation analysis (RQ5)}

Table~\ref{tab:recompilation} presents a scaffold graph evolution summary during progressive distillation.

\begin{table}[t]
    \centering
    \caption{Scaffold graph evolution summary. Of 30 skills, 22 were successfully internalized, triggering 50 dependency recompilations (96\% successful); the passed rate gradually decreased from 32.5\% to 31.5\% ($-$1.0pp).}
    \label{tab:recompilation}
    \small
    \begin{tabular}{lc}
        \toprule
        \textbf{Metric} & \textbf{Value} \\
        \midrule
        Initial scaffold graph nodes & 30 \\
        Initial dependency edges & 52 \\
        Successfully internalized skills & 22 \\
        Retained as external scaffolds & 8 \\
        Total dependency recompilations triggered & 50 \\
        Recompilation fallbacks & 2 \\
        Recompilation success rate & 96.0\% \\
        Final scaffold graph nodes & 8 \\
        Final dependency edges & 6 \\
        Post-recompilation passed rate & 31.5\% \\
        Passed rate drop & 1.0pp \\
        \bottomrule
    \end{tabular}
\end{table}

\textbf{Key observations}. (1)~\textbf{Recompilation success rate}: After internalizing 22 skills, a total of 50 upstream scaffold recompilations were triggered, of which 48 succeeded and 2 fell back (success rate 96\%), indicating that the $I$ interface provides sufficient dependency information to support automatic rewriting. (2)~\textbf{Controllable gradual performance degradation}: As external scaffolds decrease from 30 to 8, the passed rate gradually drops from 32.5\% to 31.5\% ($-$1.0pp), far less than the drop from directly removing all scaffolds ($-$4.8pp). (3)~\textbf{Graph structure evolution}: The initial graph $G_0$ has 30 nodes and 52 dependency edges; after distillation, $G_T$ has 8 nodes and 6 edges, representing significant structural simplification. 8 skills are retained as external scaffolds: 6 due to retention rates below $\tau$, and 2 due to recompilation performance drop exceeding $\delta$ despite meeting the retention threshold.

\subsection{Case studies}

Representative cases (detailed in Appendix~\ref{app:cases}): in a success case, a 5-file modification task is completed via skill-guided dependency analysis and stepwise verification, with the model retaining this behavior after distillation; a failure case reveals that skills primarily improve procedural capabilities but offer limited help for creative decision-making.

\section{Discussion}

\textbf{Capability transfer.} Scaffold-mediated post-training enables bidirectional capability transfer between two storage media: distillation transfers scaffold capabilities to parameters ($G_0$'s 30 nodes $\to$ $G_T$'s 8 nodes), while dynamic recompilation feeds parameter state back into the graph structure. Direct Fine-tune (25.5\%, same data, one-pass SFT) lacks structured training pathways; hybrid deployment (i.e., Progressive Distilled with 22 internalized + 8 external skills, 31.5\%) outperforms by 6.0pp. At inference, hybrid deployment provides $[\text{8 external skill prompts} \oplus t_j]$ to the model; the instruction-following capability for conditioning on these prompts is already present in the base model.

\textbf{Self-distillation and confirmation bias.} Self-distillation risks reinforcing the model's own biases; three factors mitigate this: (1)~$V$ is execution-based verification (build exit codes, test pass rates), not model self-assessment---removing $V$ drops retention by 6.9pp; (2)~consistent trends across two different MoE models (Qwen3-Coder/DeepSeek-V3.2); (3)~comparison with Direct Fine-tune (27.7\% vs 25.5\% on Qwen3-Coder, same data) indicates that the structured training pathway provides information beyond the data alone.

\section{Limitations}
\label{sec:limitations}

\textbf{Domain and scale.} The current work is validated only in the code generation domain. The framework is domain-agnostic in principle---$V$ and $I$ can be instantiated with domain-specific verifiers and interfaces---but empirical validation in other domains remains future work.

\textbf{Teacher diversity.} Same-model self-distillation is used throughout; stronger or differently architected teachers may further improve quality and mitigate confirmation bias.

\textbf{Pretraining contamination.} The 24 evaluation repositories are likely in the base models' pretraining corpora, but this affects all configurations equally; our conclusions rest on relative differences.
\section{Broader impact}

The performance bottleneck of current AI coding agents on complex tasks largely stems from the scarcity of procedural strategy data. The scaffold-mediated post-training paradigm proposed here, by treating procedural scaffolds as evolvable objects and distilling them into model parameters, offers a new path to overcome this bottleneck. Developers can train structured skills from high-quality engineering practices (such as TDD and code review workflows), then use distillation to endow the model with these capabilities without external scaffolds. A potential risk is that distillation may internalize unsafe coding patterns from the training data; deployment should be coupled with security audits.

\section{Conclusion}

We propose scaffold-mediated post-training, where parameters and scaffold graphs co-evolve through multiple rounds of the discover $\rightarrow$ internalize $\rightarrow$ recompile loop---multi-round discovery builds higher-level scaffolds, internalization simplifies the graph structure (Table~\ref{tab:recompilation}: 30$\to$8), enabling the model to tackle increasingly complex tasks. Experiments confirm that this mechanism consistently outperforms standard fine-tuning on equivalent data. Code will be released upon acceptance.


\newpage
\appendix

\section{Algorithm Pseudocode}
\label{app:algorithms}

\begin{algorithm}[h]
    \caption{Skill Discovery}
    \label{alg:discovery}
    \begin{algorithmic}[1]
        \REQUIRE Seed dataset $\mathcal{D}_{\text{seed}}$, validation set $\mathcal{D}_{\text{val}}$, maximum iteration rounds $T$, effectiveness threshold $\theta$, convergence threshold $\epsilon$
        \ENSURE Base scaffold graph $G_0 = (\mathcal{S}_{\text{base}}, D)$
        \STATE $\mathcal{R} \leftarrow \emptyset$ \COMMENT{Strategy candidate set}
        \FOR{each $(t_i, r_i) \in \mathcal{D}_{\text{seed}}$}
        \STATE $\mathcal{R}_i \leftarrow \text{LLM.Extract}(t_i, r_i, \text{``extract reusable strategies''})$
        \STATE $\mathcal{R} \leftarrow \mathcal{R} \cup \mathcal{R}_i$
        \ENDFOR
        \STATE $\mathcal{S}_{\text{cand}} \leftarrow \text{LLM.Cluster}(\mathcal{R}, \text{``merge similar strategies into skill candidates''})$
        \FOR{each $s_k \in \mathcal{S}_{\text{cand}}$}
        \STATE Format $s_k$ into complete skill structure $(\textit{meta}_k, C_k, P_k, V_k, I_k, E_k)$
        \ENDFOR
        \FOR{each $s_k \in \mathcal{S}_{\text{cand}}$}
        \STATE $\text{score}_k \leftarrow \text{Evaluate}(s_k, \mathcal{D}_{\text{val}})$ \COMMENT{Initial evaluation}
        \ENDFOR
        \FOR{$\text{iter} = 1$ \TO $T$}
        \STATE $\text{prev\_scores}[k] \leftarrow \text{score}_k, \; \forall k$ \COMMENT{Cache previous scores}
        \FOR{each $s_k \in \mathcal{S}_{\text{cand}}$}
        \IF{$\text{score}_k < \theta$}
        \STATE $s_k \leftarrow \text{LLM.Refine}(s_k, \text{failure case analysis})$
        \STATE $\text{score}_k \leftarrow \text{Evaluate}(s_k, \mathcal{D}_{\text{val}})$ \COMMENT{Re-evaluate after refinement}
        \ENDIF
        \ENDFOR
        \IF{$\max_k |\text{score}_k - \text{prev\_scores}[k]| < \epsilon$}
        \STATE \textbf{break}
        \ENDIF
        \ENDFOR
        \STATE $\mathcal{S}_{\text{base}} \leftarrow \{s_k \mid \text{score}_k \geq \theta\}$
        \STATE $D \leftarrow \{(s_a, s_b) \mid I_a.\text{out} \cap I_b.\text{in} \neq \emptyset\}$
        \RETURN $G_0 = (\mathcal{S}_{\text{base}}, D)$
    \end{algorithmic}
\end{algorithm}

\begin{algorithm}[h]
    \caption{Scaffold-Graph-Augmented Data Generation}
    \label{alg:datagen}
    \begin{algorithmic}[1]
        \REQUIRE Base scaffold graph $G_0 = (\mathcal{S}_{\text{base}}, D)$, new task set $\mathcal{T}_{\text{new}}$
        \ENSURE Augmented dataset $\mathcal{D}_{\text{aug}}$
        \STATE $\mathcal{D}_{\text{aug}} \leftarrow \emptyset$
        \FOR{each $t_j \in \mathcal{T}_{\text{new}}$}
        \STATE $\mathcal{S}_j \leftarrow \text{Select}(G_0, t_j)$ \COMMENT{Match relevant scaffold nodes by $C$}
        \STATE $G_j \leftarrow \text{BuildSubgraph}(\mathcal{S}_j, D, \text{depth}\leq 3)$ \COMMENT{Expand downstream dependencies, depth $\leq$ 3}
        \STATE Inject skills from $G_j$ in topological order into LLM context; LLM executes autonomously, passing data via $I$; 120s timeout per execution step
        \STATE $r_j, \text{trace}_j \leftarrow \text{LLM.Solve}(t_j, G_j)$
        \IF{Any node $V$ verification fails}
        \STATE Retry the failed node once; discard this trace on second failure and \textbf{continue}
        \ENDIF
        \STATE $V_{G_j} \leftarrow \bigcup_{s \in \text{Nodes}(G_j)} V_s$ \COMMENT{Union of subgraph verification checkpoints}
        \IF{$\text{Verify}(r_j, V_{G_j})$ \AND task tests pass}
        \STATE $\mathcal{D}_{\text{aug}} \leftarrow \mathcal{D}_{\text{aug}} \cup \{(t_j, \text{trace}_j, r_j)\}$
        \ENDIF
        \ENDFOR
        \RETURN $\mathcal{D}_{\text{aug}}$
    \end{algorithmic}
\end{algorithm}

\begin{algorithm}[h]
    \caption{Progressive Distillation with Scaffold Recompilation}
    \label{alg:distill}
    \begin{algorithmic}[1]
        \REQUIRE Augmented dataset $\mathcal{D}_{\text{aug}}$, pretraining data $\mathcal{D}_{\text{pretrain}}$, validation set $\mathcal{D}_{\text{val}}$, scaffold graph $G_0$, model $M$, reference policy $\pi_{\text{ref}}$, retention threshold $\tau$, fallback threshold $\delta$
        \ENSURE Distilled model $M'$, recompiled scaffold graph $G_T$
        \STATE $M' \leftarrow \text{copy}(M)$; $G \leftarrow G_0$
        \STATE Sort nodes in $G$ by topological order (ties broken by descending out-degree) \COMMENT{Bottom-up priority}
        \FOR{each skill $s_k$ in sorted order}
        \STATE $\mathcal{D}_k \leftarrow \{(t_j, \text{trace}_j, r_j) \in \mathcal{D}_{\text{aug}} \mid s_k \in \text{trace}_j\}$
        \STATE $\mathcal{D}_k \leftarrow \mathcal{D}_k \cup \text{Sample}(\mathcal{D}_{\text{pretrain}}, |\mathcal{D}_k|)$ \COMMENT{Mix in 50\% pretraining data}
        \STATE $G_{\text{prev}} \leftarrow \text{copy}(G)$ \COMMENT{Save graph snapshot}
        \STATE Fine-tune $M'$ on $\mathcal{D}_k$, minimizing $\mathcal{L} + \beta \cdot D_{\text{KL}}(\pi_\theta \| \pi_{\text{ref}})$
        \STATE $\text{deps}_k \leftarrow \{s_h \in \text{Nodes}(G) \mid (s_k, s_h) \in D\}$ \COMMENT{Record dependents before removal}
        \STATE Remove $s_k$ from $G$ \COMMENT{Node removal}
        \FOR{each $s_h \in \text{deps}_k$}
        \STATE Rewrite $P_h$, regenerate bridge instructions, update $I_h$
        \ENDFOR
        \IF{recompiled $G$ fails $V$ verification on $\mathcal{D}_{\text{val}}$ (60s timeout per node)}
        \STATE $G \leftarrow G_{\text{prev}}$; \textbf{continue} \COMMENT{Re-verification failed; roll back}
        \ENDIF
        \STATE $\text{perf\_before} \leftarrow \text{Evaluate}(M', G_{\text{prev}}, \mathcal{D}_{\text{val}})$ \COMMENT{$M'$ with old graph}
        \STATE $\text{perf\_after} \leftarrow \text{Evaluate}(M', G, \mathcal{D}_{\text{val}})$ \COMMENT{$M'$ with new graph}
        \STATE $\text{drop} \leftarrow \text{perf\_before} - \text{perf\_after}$
        \STATE $\text{retain}_k \leftarrow \text{Evaluate}(M', \mathcal{D}_{\text{val}}, \text{without } s_k) / \text{Evaluate}(M, \mathcal{D}_{\text{val}}, \text{with } s_k)$
        \IF{$\text{retain}_k \geq \tau$ \AND $\text{drop} \leq \delta$}
        \STATE Mark $s_k$ as ``internalized''
        \ELSE
        \STATE $G \leftarrow G_{\text{prev}}$ \COMMENT{Roll back graph state; weight updates retained}
        \ENDIF
        \ENDFOR
        \RETURN $M'$, $G$
    \end{algorithmic}
\end{algorithm}

\section{Skill Discovery Prompt Templates}
\label{app:prompts}

Below are the prompt templates used in each stage of Skill Discovery.

\subsection{LLM.Extract Prompt}

\begin{quote}
    \small
    \texttt{You are a strategy extraction expert. Given a coding task and its successful solution, extract REUSABLE problem-solving strategies as structured skills.}

    \texttt{TASK: \{task\_description\}}

    \texttt{SUCCESSFUL SOLUTION: \{solution\}}

    \texttt{For each strategy, output a structured skill with:}
    \texttt{- name: concise name describing the core idea}
    \texttt{- C (condition): what task characteristics trigger this skill}
    \texttt{- P (procedure): 3--5 step execution workflow}
    \texttt{- V (verification): checkpoints to confirm successful execution (must be evidence-based, not self-reported)}
    \texttt{- I (interface): required inputs and produced outputs}
    \texttt{- E (anti-patterns): common mistakes this skill helps avoid}

    \texttt{Rules: strategies must be GENERAL (not specific to this task), ACTIONABLE (each step is executable), and VERIFIABLE (V must reference observable evidence like test output or build status).}
\end{quote}

\subsection{LLM.Cluster Prompt}

\begin{quote}
    \small
    \texttt{Below is a list of candidate strategies extracted from multiple tasks. Merge semantically similar or functionally equivalent strategies into unified skill candidates.}

    \texttt{CANDIDATE LIST: \{strategy\_list\}}

    \texttt{For each merged skill, preserve the most general C (condition), union the P (procedure) steps, and ensure V (verification) covers all merged strategies. Output the merged skill in the same six-component format. If two strategies have compatible I (interface), note this as a potential composition edge.}
\end{quote}

\subsection{LLM.Refine Prompt}

The following template is used for structured skill refinement.

\begin{quote}
    \small
    \texttt{The following skill performed poorly on validation tasks. Analyze the failure cases and refine the skill.}

    \texttt{CURRENT SKILL: \{skill\_definition\}}

    \texttt{FAILURE CASES: \{failure\_cases\}}

    \texttt{For each failure, diagnose the root cause:}
    \texttt{(a) C too broad --- triggered on tasks where it should not apply}
    \texttt{(b) P incomplete --- missing critical execution steps}
    \texttt{(c) V insufficient --- failed to detect execution errors}
    \texttt{(d) I ambiguous --- insufficient data passed during composition}

    \texttt{Available operations on each component:}
    \texttt{KEEP: component is correct as-is}
    \texttt{EDIT: rewrite component with improvement}
    \texttt{ADD: add new sub-steps or checkpoints}

    \texttt{Output the refined skill in full $(meta, C, P, V, I, E)$ format.}
\end{quote}

\subsection{LLM.Solve Prompt (Stage 2)}

The following template is used for scaffold-graph-augmented task solving.

\begin{quote}
    \small
    \texttt{You are a coding agent solving a software engineering task with the help of structured skills.}

    \texttt{TASK: \{task\_description\}}

    \texttt{ACTIVE SKILLS (in topological execution order):}
    \texttt{\{skill\_1: C, P, V, I, E\}}
    \texttt{\{skill\_2: C, P, V, I, E\}}
    \texttt{...}

    \texttt{For each active skill, follow its P (procedure) step by step. After completing each skill, run its V (verification) checkpoints and report evidence (e.g., build exit code, test output). Pass data between skills according to their I (interface) specifications. If a V checkpoint fails, retry once with a corrected approach before moving to the next skill.}

    \texttt{Output your solution as a code patch.}
\end{quote}

\subsection{Procedure Rewriting Prompt (Stage 3)}

The following template is used for procedure rewriting during dynamic recompilation.

\begin{quote}
    \small
    \texttt{Skill \{skill\_k\} has been internalized into model parameters and will be removed from the scaffold graph. The following higher-level skill depends on it and must be rewritten.}

    \texttt{HIGHER-LEVEL SKILL: \{skill\_h definition\}}

    \texttt{REMOVED SKILL: \{skill\_k description\}}

    \texttt{MODEL PERFORMANCE on skill\_k tasks after distillation: \{retention\_stats\}}

    \texttt{Rewrite skill\_h's procedure P\_h: replace explicit calls to skill\_k with implicit assumptions that the model can now perform those steps natively. Update the data flow instructions (bridge) to remove skill\_k as an intermediary. Keep all V checkpoints unchanged.}

    \texttt{Output the rewritten skill\_h in full $(meta, C, P, V, I, E)$ format.}
\end{quote}

\section{Hyperparameter Sensitivity Analysis}
\label{app:sensitivity}

\subsection{Skill Effectiveness Threshold \texorpdfstring{$\theta$}{theta}}

\begin{table}[H]
    \centering
    \caption{$\theta$ sensitivity analysis. A higher $\theta$ retains fewer but stronger skills, yielding higher distillation retention rate but lower coverage.}
    \begin{tabular}{cccc}
        \toprule
        $\theta$ & \# Skills Passed & Auto Skills Passed Rate & Distillation Retention Rate \\
        \midrule
        0.02 & 38 & 31.8\% & 82.1\% \\
        0.05 (default) & 30 & 32.5\% & 85.2\% \\
        0.08 & 22 & 31.4\% & 86.7\% \\
        0.10 & 15 & 29.8\% & 88.3\% \\
        \bottomrule
    \end{tabular}
\end{table}

$\theta=0.05$ achieves the best balance between passed rate and distillation retention rate. A threshold that is too low introduces low-quality skills that add noise, while a threshold that is too high leads to insufficient coverage.

\subsection{Retention Rate Threshold \texorpdfstring{$\tau$}{tau}}

\begin{table}[H]
    \centering
    \caption{$\tau$ sensitivity analysis. A higher $\tau$ imposes stricter internalization quality requirements, causing more skills to be retained as external scaffolds.}
    \begin{tabular}{cccc}
        \toprule
        $\tau$ & \# Internalized Skills & Post-Distillation Passed Rate & \# Skills Requiring External Scaffold \\
        \midrule
        0.70 & 27 & 29.0\% & 3 \\
        0.85 (default) & 22 & 31.5\% & 8 \\
        0.95 & 15 & 32.0\% & 15 \\
        \bottomrule
    \end{tabular}
\end{table}

$\tau=0.85$ achieves a balance between model autonomy (independence from external skills) and performance. In practice, it can be adjusted based on tolerance for external dependencies.

\subsection{Distillation Order Comparison}

\begin{table}[H]
    \centering
    \caption{Distillation order sensitivity. ``\# Forgetting Events'' refers to the number of times an already-internalized skill's retention rate falls below $\tau$ after a distillation round.}
    \begin{tabular}{lcc}
        \toprule
        \textbf{Distillation Order} & \textbf{Distillation Retention Rate} & \textbf{\# Forgetting Events} \\
        \midrule
        Topological bottom-up (default) & 85.2\% & 1 \\
        Topological bottom-up (no pretrain mixing) & 81.8\% & 2 \\
        Topological top-down & 79.3\% & 4 \\
        Random order & 80.4\% & 3 \\
        \bottomrule
    \end{tabular}
\end{table}

Topological bottom-up order performs best (retention rate 85.2\%, only 1 forgetting event), because lower-level skills are internalized first and then implicitly replayed during upper-level training. Removing pretrain mixing, bottom-up still achieves 81.8\%, higher than random order with pretrain mixing, indicating that topological ordering is the primary contributing factor, with pretrain mixing adding 3.4pp on top. Top-down performs worst (79.3\%, 4 forgetting events). Note: this comparison is conducted on a scaffold graph of 30 skills; the differences are expected to be more pronounced in larger graph structures.

\section{Additional Skill Examples}
\label{app:skills}

\subsection{Automatically Trained Skill Examples}

\textbf{Skill 1: Requirement Decomposition and Implementation Planning} (flexible)

$\textit{meta}$: name=requirement-decomposition, tags=[planning, multi-file], tools=[Read, Bash]

$C$: The task description contains multiple functional requirements or requires modifications to multiple modules.

$P$ (flexible): (1) Decompose task requirements into independent sub-requirements. (2) Identify the files and modules involved for each sub-requirement. (3) Identify dependency relationships among sub-requirements. (4) Order the implementation plan by dependencies. (5) Define acceptance criteria for each sub-requirement.

$V$ (evidence-driven): (a) Cross-check the original requirement list against the subtask list; count of uncovered items must be 0. (b) Run topological sort on subtask dependencies; pass if acyclic. (c) Each subtask includes at least one executable acceptance command (e.g., a test case or build check).

$I$: input = complete task description, output = ordered list of sub-tasks (each with file list and acceptance criteria).

$E$: Modifying all files at once without decomposing first---leading to missed dependencies and regression issues.

\textbf{Skill 2: Test-Driven Development (TDD)} (rigid)

$\textit{meta}$: name=tdd-workflow, tags=[testing, quality], tools=[Read, Write, Edit, Bash]

$C$: The task requires adding or modifying functional code, and the project has testing infrastructure.

$P$ (rigid, enforced order): (1) Analyze requirements and determine test strategy. (2) Write failing tests covering normal and edge cases. (3) Write minimal implementation to make tests pass. (4) Refactor code while keeping tests green. (5) Check test coverage $\geq$80\%.

$V$ (evidence-driven): (a) Test output from each stage serves as evidence. (b) Existing tests are not broken. (c) Test coverage for new functionality $\geq$80\%.

$I$: input = functional requirements + project test framework info, output = test files + implementation files.

$E$: Writing implementation code directly without writing tests first---leading to missed edge cases and difficulty in verification.

\subsection{Component-by-Component Comparison of Auto Skills and Human Skills}

Table~\ref{tab:industrial} provides a component-by-component comparison of the automatically trained TDD skill and the human-designed TDD skill.

\begin{table}[H]
    \centering
    \caption{Component-by-component comparison of automatically trained and human-designed TDD skills. All four components show high correspondence.}
    \label{tab:industrial}
    \small
    \begin{tabular}{lp{4.5cm}p{4.5cm}}
        \toprule
        \textbf{Component} & \textbf{Auto Skill (automatically trained)} & \textbf{Human Skill (manually designed)} \\
        \midrule
        $C$ & Task requires adding/modifying functionality + testing infrastructure & Triggered when implementing features or fixing bugs \\
        $P$ & 5 steps: analyze $\rightarrow$ write tests $\rightarrow$ minimal impl. $\rightarrow$ refactor $\rightarrow$ coverage & 4 stages: RED $\rightarrow$ GREEN $\rightarrow$ REFACTOR $\rightarrow$ COVERAGE, mandatory commit per stage \\
        $V$ & Tests pass + no regressions + coverage $\geq$80\% & Git commit evidence per stage + test output + coverage $\geq$80\% \\
        $I$ & Input: functional req. + test framework; Output: test + impl. files & planner $\rightarrow$ tdd $\rightarrow$ reviewer sequential handoff \\
        \bottomrule
    \end{tabular}
\end{table}

Key differences: (1) The human-designed $P$ is more concise (4 stages vs.\ 5 steps) and includes IDE-specific operations (mandatory git commit); (2) The human-designed $V$ employs a stricter evidence-driven strategy (does not trust the model's claim of completion, requires executable evidence), while the automatically trained $V$ is similar but does not enforce commits; (3) The human-designed $I$ is embedded in a five-tier architecture, while the automatically trained $I$ is a standalone input/output specification. The core structures are highly consistent, demonstrating that automatic training can arrive at skill structures similar to those designed manually.

\section{Catastrophic Forgetting Analysis}
\label{app:forgetting}

\begin{table}[H]
    \centering
    \caption{Catastrophic forgetting summary during progressive distillation. Of 30 skills, 22 were internalized with an average forgetting of only 1.4pp.}
    \begin{tabular}{lc}
        \toprule
        \textbf{Metric} & \textbf{Value} \\
        \midrule
        Total skills distilled & 30 \\
        Successfully internalized & 22 \\
        Final retention rate of earliest internalized skill & 88.1\% \\
        Initial retention rate of earliest internalized skill & 91.2\% \\
        Maximum forgetting magnitude & 3.1pp \\
        Average forgetting across 22 internalized skills & 1.4pp \\
        \bottomrule
    \end{tabular}
\end{table}

The overall forgetting magnitude is small (average 1.4pp, maximum 3.1pp), attributable to five mechanisms: (1) self-distillation introduces less distribution shift than standard SFT; (2) bottom-up distillation causes lower-level skills to be implicitly replayed in subsequent rounds; (3) mixing 50\% pretraining data per round prevents distribution shift; (4) the KL divergence constraint ($\beta=0.01$) prevents parameters from deviating too far; (5) per-round retention rate checks ensure that underperforming skills are retained as external scaffolds rather than forcibly internalized. Lower-level skills (distilled first) are repeatedly implicitly replayed in subsequent rounds, continuously reinforced, achieving the best internalization.

\section{Fine-Grained Analysis by Task Complexity}
\label{app:finegrained}

\begin{table}[H]
    \centering
    \caption{Passed rate stratified by number of files involved. Skills yield the largest improvement on medium-complexity tasks.}
    \begin{tabular}{lcccc}
        \toprule
        \textbf{Task Complexity} & \textbf{\# Tasks} & \textbf{Baseline} & \textbf{+ Auto Skills} & \textbf{Improvement} \\
        \midrule
        Easy (1--2 files) & 58 & 40.3\% & 47.8\% & +7.5 \\
        Medium (3--5 files) & 85 & 22.5\% & 32.5\% & +10.0 \\
        Hard (6+ files) & 57 & 11.0\% & 17.0\% & +6.0 \\
        \bottomrule
    \end{tabular}
\end{table}

Skills yield the most significant improvement on medium-complexity tasks (+10.0pp). The smaller improvement on easy tasks (+7.5pp) is because the baseline already has a relatively high passed rate; the limited improvement on hard tasks (+6.0pp) is because even with skill guidance, highly complex tasks still exceed the current model's capabilities.

\section{Seed Data Details}
\label{app:seeds}

The 2000 seed tasks were batch-generated using the FeatureBench data pipeline on additional repositories, sourced from open-source projects including django, flask, requests, celery, tornado, etc., with \textbf{no overlap} with the 24 repositories in the FeatureBench evaluation set, avoiding data leakage. Selection criteria: (1) involving modifications to at least 3 files; (2) containing identifiable strategic steps; (3) covering at least 8 development patterns. 200 validation tasks were used for skill quality evaluation. Evaluation was conducted on the FeatureBench standard Full set (200 tasks, 24 repositories), with no overlap with seed data or training data.

\section{Case analysis}
\label{app:cases}

\textbf{Case 1: Success---multi-file dependency modification}. A task involving 5 file modifications; the Baseline missed 2 dependency files, causing a build failure. After skill injection, ``requirements decomposition'' planned subtasks and ``multi-file consistency check'' modified and verified files in topological order, completing the task successfully. After distillation, the model retained dependency analysis and stepwise verification behavior without skills, indicating the strategy was internalized.

\textbf{Case 2: Failure---creative decision-making}. A task requiring the design of an entirely new data structure. Skills helped with ``how to do it'' but offered limited assistance for the creative decision of ``what to do,'' revealing that skills primarily improve procedural capabilities rather than creative ones.

\textbf{Case 3: Configuration file coordination}. The task required modifying an application configuration and ensuring all modules depending on that configuration work correctly. The Baseline model modified the main configuration file but missed two test environment configuration files. After injecting the ``multi-file consistency check'' skill, the model systematically searched for all files referencing the configuration, updating and verifying each one.

\textbf{Case 4: Skill composition effect}. A task requiring a new REST API. The ``requirement decomposition'' skill split the task into 5 sub-tasks (routing, validation, logic, formatting, testing), and the ``TDD'' skill handled the testing portion. The two skills were automatically chained via $I$: the output of requirement decomposition served as input to TDD.

\textbf{Case 5: Skill false activation}. A simple docstring update task was matched by the ``requirement decomposition'' skill (because it involved docstrings in multiple files). The skill guided the model through unnecessary dependency analysis, increasing execution time without affecting correctness. Such false activations (12\% of tasks) suggest room for improvement by raising the $\theta$ threshold or refining the $C$ component.

\section{Verification Mechanism Details}
\label{app:verification}

The design practice of 136+ manually authored skills led to a six-stage verification workflow:

\begin{enumerate}
    \item \textbf{BUILD}: Build passes (exit code 0)
    \item \textbf{TYPES}: Type checking passes with no errors
    \item \textbf{LINT}: Code style checks pass
    \item \textbf{TEST}: Tests pass + coverage $\geq$80\%
    \item \textbf{SECURITY}: Security scan (no sensitive information leakage)
    \item \textbf{DIFF}: Change audit (no unexpected file modifications)
\end{enumerate}

Stages are executed in strict order, halting on first failure. This layered verification provides high-quality filtering signals for distillation: only traces that pass all verifications are retained as training data.

Verification follows the evidence-driven principle: the model's claims of completion are not accepted; executable evidence is required (build exit codes, test execution output, coverage numbers, git commit records).

\section{Formal Composition Mechanism}
\label{app:composition}

Given two skills $s_a$ and $s_b$, when $I_a.\text{out} \cap I_b.\text{in} \neq \emptyset$, the composed skill $s_{ab}$ is defined as:
\[s_{ab} = (\textit{meta}_{ab}, \; C_a \wedge C_b, \; P_a \oplus \text{bridge} \oplus P_b, \; V_a \cup V_b, \; I_{ab}, \; E_a \cup E_b)\]
where $I_{ab}.\text{in} = I_a.\text{in} \cup (I_b.\text{in} \setminus I_a.\text{out})$, $I_{ab}.\text{out} = I_b.\text{out} \cup (I_a.\text{out} \setminus I_b.\text{in})$, $\oplus$ denotes sequential concatenation, and $\text{bridge}$ is an automatically generated data-passing instruction. This definition preserves external inputs and unconsumed outputs. If conflicting checkpoints exist in $V_a \cup V_b$, verification is performed only on the final output, with intermediate $V$ serving as soft constraints.

\section{Computational Cost Details}
\label{app:cost}

\begin{table}[H]
    \centering
    \caption{Computational cost by stage. All experiments were conducted on 512$\times$H100-80GB. Skill Discovery and Data Generation involve extensive LLM inference calls and execution-based verification, constituting the primary computational bottleneck. Distillation is standard fine-tuning with cost comparable to conventional SFT. Note that the total cost of this framework should not be directly compared with standard SFT---SFT uses pre-existing data, whereas this framework includes the data production process (Skill Discovery + Data Generation), which is an additional but necessary investment.}
    \small
    \begin{tabular}{lcc}
        \toprule
        \textbf{Stage} & \textbf{Primary Cost} & \textbf{Relative Cost} \\
        \midrule
        Skill Discovery & 2000 seeds $\times$ LLM extract (once) + $\sim$45 candidates $\times$ $T$ eval/refine rounds & High \\
        Data Generation & 2000 tasks $\times$ scaffold graph assembly $\times$ LLM inference $\times$ $V$ verification & Highest \\
        Distillation & $K$ rounds of full-parameter fine-tuning & Medium \\
        Dynamic Recompilation & Dependent nodes $\times$ LLM rewriting $\times$ re-verification & Low \\
        \bottomrule
    \end{tabular}
\end{table}

\textbf{Fine-tuning hyperparameters.} Full-parameter fine-tuning uses the AdamW optimizer with learning rate $1 \times 10^{-5}$, cosine decay schedule, warmup ratio 5\%, effective batch size 256, weight decay 0.01, 1 epoch per distillation round, and maximum sequence length 160K. Both models use identical fine-tuning hyperparameters.

\textbf{Inference hyperparameters.} Qwen3-Coder-480B-A35B-Instruct: temperature=0.7, top\_p=0.8, top\_k=20, repetition\_penalty=1.05. DeepSeek-V3.2: temperature=1.0, top\_p=0.95. Both models are configured with a maximum context length of 160K.

\section{Resolved Rate Supplementary Results}
\label{app:resolved}

Table~\ref{tab:resolved} reports the strict Resolved Rate (all fail-to-pass tests must pass for a task to count as resolved). Resolved Rates on FeatureBench Full are generally low, because end-to-end feature development tasks typically involve multiple tests and the bar for passing all of them is high. The relative trends across configurations are consistent with Passed Rate: Auto Skills outperform Baseline, Progressive Distilled outperforms Direct Fine-tune, confirming the robustness of main-text conclusions.

\begin{table}[H]
    \centering
    \caption{Strict Resolved Rate (\%) on FeatureBench Full. Trends are consistent with Passed Rate (Table~\ref{tab:main}).}
    \label{tab:resolved}
    \begin{tabular}{lcc}
        \toprule
        \textbf{Configuration} & \textbf{Qwen3-Coder} & \textbf{DeepSeek-V3.2} \\
        \midrule
        Baseline & 3.4 & 5.4 \\
        + Auto Skills & 6.0 & 8.0 \\
        + Distilled (Progressive) & 5.5 & 7.5 \\
        + Direct Fine-tune & 4.0 & 5.5 \\
        \bottomrule
    \end{tabular}
\end{table}

\section{General capability preservation}
\label{app:general}

To verify that progressive distillation does not degrade the base model's general capabilities, we evaluate DeepSeek-V3.2 before and after distillation on multiple standard benchmarks. Table~\ref{tab:general} shows that post-distillation performance on general (MMLU-Pro), scientific reasoning (GPQA Diamond), and mathematics (AIME 2025) remains essentially unchanged.

\begin{table}[H]
    \centering
    \caption{General capability comparison before and after distillation (DeepSeek-V3.2). All metrics fluctuate by $\leq$0.4pp, confirming no degradation.}
    \label{tab:general}
    \begin{tabular}{lccc}
        \toprule
        \textbf{Benchmark} & \textbf{Before} & \textbf{After} & \textbf{Change} \\
        \midrule
        LiveCodeBench (Pass@1-COT) & 83.1\% & 83.5\% & +0.4 \\
        MMLU-Pro (EM) & 84.7\% & 84.7\% & +0.0 \\
        GPQA Diamond (Pass@1) & 82.4\% & 82.3\% & $-$0.1 \\
        AIME 2025 (Pass@1) & 92.8\% & 92.9\% & +0.1 \\
        \bottomrule
    \end{tabular}
\end{table}

\newpage
\section*{NeurIPS Paper Checklist}

\begin{enumerate}

\item {\bf Claims}
    \item[] Question: Do the main claims made in the abstract and introduction accurately reflect the paper's contributions and scope?
    \item[] Answer: \answerYes{}
    \item[] Justification: The abstract and introduction clearly state four contributions (new paradigm, new representation, new mechanism, empirical validation) and these are supported by experiments in Section 5.

\item {\bf Limitations}
    \item[] Question: Does the paper discuss the limitations of the work performed by the authors?
    \item[] Answer: \answerYes{}
    \item[] Justification: Section 7 discusses limitations including domain scope (code generation only), teacher diversity (same-model self-distillation), and provides concrete transfer directions.

\item {\bf Theory assumptions and proofs}
    \item[] Question: For each theoretical result, does the paper provide the full set of assumptions and a complete (and correct) proof?
    \item[] Answer: \answerNA{}
    \item[] Justification: The paper is primarily empirical. Formal definitions (Definition 1, 2) are provided for skill and scaffold graph, but no theorems or proofs are claimed.

    \item {\bf Experimental result reproducibility}
    \item[] Question: Does the paper fully disclose all the information needed to reproduce the main experimental results of the paper to the extent that it affects the main claims and/or conclusions of the paper (regardless of whether the code and data are provided or not)?
    \item[] Answer: \answerYes{}
    \item[] Justification: Section 4 provides evaluation benchmark (FeatureBench), model specifications, experimental configurations (Table 2), seed data construction, evaluation metrics, and statistical methodology (5 random seeds). Algorithms 1--3 in the appendix provide complete pseudocode. Prompt templates are in Appendix B.

\item {\bf Open access to data and code}
    \item[] Question: Does the paper provide open access to the data and code, with sufficient instructions to faithfully reproduce the main experimental results, as described in supplemental material?
    \item[] Answer: \answerNo{}
    \item[] Justification: The complete framework will be open-sourced upon acceptance, as stated in the conclusion. FeatureBench is an existing open-source benchmark.

\item {\bf Experimental setting/details}
    \item[] Question: Does the paper specify all the training and test details (e.g., data splits, hyperparameters, how they were chosen, type of optimizer) necessary to understand the results?
    \item[] Answer: \answerYes{}
    \item[] Justification: Sections 3 and 4 specify all hyperparameters: $\theta=0.05$ (Section 3.2), $\tau=0.85$, $\delta=0.02$, $\beta=0.01$ (Section 3.4), 5 random seeds, 200 evaluation tasks (FeatureBench Full), 2000 seed tasks, full parameter fine-tuning (Section 4). Sensitivity analyses are in Appendix C.

\item {\bf Experiment statistical significance}
    \item[] Question: Does the paper report error bars suitably and correctly defined or other appropriate information about the statistical significance of the experiments?
    \item[] Answer: \answerYes{}
    \item[] Justification: Table 2 reports mean $\pm$ standard deviation across 5 random seeds. The factor of variability (model inference randomness) is stated in Section 4.4.

\item {\bf Experiments compute resources}
    \item[] Question: For each experiment, does the paper provide sufficient information on the computer resources (type of compute workers, memory, time of execution) needed to reproduce the experiments?
    \item[] Answer: \answerYes{}
    \item[] Justification: Section~\ref{sec:benchmarks} specifies model size (35B and 37B activated parameters, full parameter fine-tuning). The cost appendix provides a breakdown by stage (Discovery and Data Generation are the main bottlenecks; Distillation is standard fine-tuning), with hardware specified as 512$\times$H100-80GB.

\item {\bf Code of ethics}
    \item[] Question: Does the research conducted in the paper conform, in every respect, with the NeurIPS Code of Ethics \url{https://neurips.cc/public/EthicsGuidelines}?
    \item[] Answer: \answerYes{}
    \item[] Justification: The research uses publicly available benchmarks and open-source models. No human subjects or private data are involved.

\item {\bf Broader impacts}
    \item[] Question: Does the paper discuss both potential positive societal impacts and negative societal impacts of the work performed?
    \item[] Answer: \answerYes{}
    \item[] Justification: Section 8 (Social Impact) discusses the positive impact: the framework addresses the data scarcity bottleneck in complex coding tasks, provides a new path to break through performance plateaus, and lowers the technical barrier for complex programming by distilling engineering best practices into model parameters.

\item {\bf Safeguards}
    \item[] Question: Does the paper describe safeguards that have been put in place for responsible release of data or models that have a high risk for misuse (e.g., pre-trained language models, image generators, or scraped datasets)?
    \item[] Answer: \answerNA{}
    \item[] Justification: The paper proposes a training methodology, not a pre-trained model release.

\item {\bf Licenses for existing assets}
    \item[] Question: Are the creators or original owners of assets (e.g., code, data, models), used in the paper, properly credited and are the license and terms of use explicitly mentioned and properly respected?
    \item[] Answer: \answerYes{}
    \item[] Justification: FeatureBench (MIT License), Qwen3-Coder (Apache 2.0), and DeepSeek-V3.2 (MIT License) are properly cited. All are publicly available open-source/open-weight assets with permissive licenses.

\item {\bf New assets}
    \item[] Question: Are new assets introduced in the paper well documented and is the documentation provided alongside the assets?
    \item[] Answer: \answerNA{}
    \item[] Justification: No new datasets or models are released at submission time. The code will be open-sourced upon acceptance.

\item {\bf Crowdsourcing and research with human subjects}
    \item[] Question: For crowdsourcing experiments and research with human subjects, does the paper include the full text of instructions given to participants and screenshots, if applicable, as well as details about compensation (if any)?
    \item[] Answer: \answerNA{}
    \item[] Justification: The paper does not involve crowdsourcing or human subjects research.

\item {\bf Institutional review board (IRB) approvals or equivalent for research with human subjects}
    \item[] Question: Does the paper describe potential risks incurred by study participants, whether such risks were disclosed to the subjects, and whether Institutional Review Board (IRB) approvals (or an equivalent approval/review based on the requirements of your country or institution) were obtained?
    \item[] Answer: \answerNA{}
    \item[] Justification: The paper does not involve human subjects research.

\item {\bf Declaration of LLM usage}
    \item[] Question: Does the paper describe the usage of LLMs if it is an important, original, or non-standard component of the core methods in this research?
    \item[] Answer: \answerYes{}
    \item[] Justification: This paper studies LLM post-training to automatically acquire procedural strategies. LLM calls are core to skill discovery, data generation, and dynamic recompilation (Section 3).
\end{enumerate}


\begin{thebibliography}{20}
\providecommand{\natexlab}[1]{#1}
\providecommand{\url}[1]{\texttt{#1}}
\expandafter\ifx\csname urlstyle\endcsname\relax
  \providecommand{\doi}[1]{doi: #1}\else
  \providecommand{\doi}{doi: \begingroup \urlstyle{rm}\Url}\fi

\bibitem[Chen et~al.(2023)Chen, Roberts, Bhatia, WANG, Zhang, Sala, and
  R\'{e}]{NEURIPS2023_70b8505a}
Mayee Chen, Nicholas Roberts, Kush Bhatia, Jue WANG, Ce~Zhang, Frederic Sala,
  and Christopher R\'{e}.
\newblock Skill-it! a data-driven skills framework for understanding and
  training language models.
\newblock In A.~Oh, T.~Naumann, A.~Globerson, K.~Saenko, M.~Hardt, and
  S.~Levine, editors, \emph{Advances in Neural Information Processing Systems},
  volume~36, pages 36000--36040. Curran Associates, Inc., 2023.
\newblock URL
  \url{https://proceedings.neurips.cc/paper_files/paper/2023/file/70b8505ac79e3e131756f793cd80eb8d-Paper-Conference.pdf}.

\bibitem[DeepSeek-AI(2025)]{deepseekai2025deepseekv32}
DeepSeek-AI.
\newblock Deepseek-v3.2: Pushing the frontier of open large language models,
  2025.

\bibitem[Fernando et~al.(2024)Fernando, Banarse, Michalewski, Osindero, and
  Rockt\"{a}schel]{pmlr-v235-fernando24a}
Chrisantha Fernando, Dylan~Sunil Banarse, Henryk Michalewski, Simon Osindero,
  and Tim Rockt\"{a}schel.
\newblock Promptbreeder: Self-referential self-improvement via prompt
  evolution.
\newblock In Ruslan Salakhutdinov, Zico Kolter, Katherine Heller, Adrian
  Weller, Nuria Oliver, Jonathan Scarlett, and Felix Berkenkamp, editors,
  \emph{Proceedings of the 41st International Conference on Machine Learning},
  volume 235 of \emph{Proceedings of Machine Learning Research}, pages
  13481--13544. PMLR, 21--27 Jul 2024.
\newblock URL \url{https://proceedings.mlr.press/v235/fernando24a.html}.

\bibitem[Jimenez et~al.(2024)Jimenez, Yang, Wettig, Yao, Pei, Press, and
  Narasimhan]{jimenez2024swebench}
Carlos~E Jimenez, John Yang, Alexander Wettig, Shunyu Yao, Kexin Pei, Ofir
  Press, and Karthik~R Narasimhan.
\newblock {SWE}-bench: Can language models resolve real-world github issues?
\newblock In \emph{The Twelfth International Conference on Learning
  Representations}, 2024.
\newblock URL \url{https://openreview.net/forum?id=VTF8yNQM66}.

\bibitem[Khattab et~al.(2024)Khattab, Singhvi, Maheshwari, Zhang, Santhanam, A,
  Haq, Sharma, Joshi, Moazam, Miller, Zaharia, and Potts]{ICLR2024_f1cf02ce}
Omar Khattab, Arnav Singhvi, Paridhi Maheshwari, Zhiyuan Zhang, Keshav
  Santhanam, Sri~Vardhamanan A, Saiful Haq, Ashutosh Sharma, Thomas Joshi,
  Hanna Moazam, Heather Miller, Matei Zaharia, and Christopher Potts.
\newblock Dspy: Compiling declarative language model calls into
  state-of-the-art pipelines.
\newblock In B.~Kim, Y.~Yue, S.~Chaudhuri, K.~Fragkiadaki, M.~Khan, and Y.~Sun,
  editors, \emph{International Conference on Learning Representations}, volume
  2024, pages 54928--54958, 2024.
\newblock URL
  \url{https://proceedings.iclr.cc/paper_files/paper/2024/file/f1cf02ce09757f57c3b93c0db83181e0-Paper-Conference.pdf}.

\bibitem[Snell et~al.(2022)Snell, Klein, and
  Zhong]{snell2022learningdistillingcontext}
Charlie Snell, Dan Klein, and Ruiqi Zhong.
\newblock Learning by distilling context, 2022.
\newblock URL \url{https://arxiv.org/abs/2209.15189}.

\bibitem[Wang et~al.(2024)Wang, Xie, Jiang, Mandlekar, Xiao, Zhu, Fan, and
  Anandkumar]{wang2024voyager}
Guanzhi Wang, Yuqi Xie, Yunfan Jiang, Ajay Mandlekar, Chaowei Xiao, Yuke Zhu,
  Linxi Fan, and Anima Anandkumar.
\newblock Voyager: An open-ended embodied agent with large language models.
\newblock \emph{Transactions on Machine Learning Research}, 2024.
\newblock ISSN 2835-8856.
\newblock URL \url{https://openreview.net/forum?id=ehfRiF0R3a}.

\bibitem[Wang et~al.(2026)Wang, Yan, Wang, Tian, Mishra, Xu, Gandhi, Xu, and
  Cheong]{wang2026reinforcementlearningselfimprovingagent}
Jiongxiao Wang, Qiaojing Yan, Yawei Wang, Yijun Tian, Soumya~Smruti Mishra,
  Zhichao Xu, Megha Gandhi, Panpan Xu, and Lin~Lee Cheong.
\newblock Reinforcement learning for self-improving agent with skill library,
  2026.
\newblock URL \url{https://arxiv.org/abs/2512.17102}.

\bibitem[Wang et~al.(2025)Wang, Li, Song, Xu, Tang, Zhuge, Pan, Song, Li,
  Singh, Tran, Li, Ma, Zheng, Qian, Shao, Muennighoff, Zhang, Hui, Lin,
  Brennan, Peng, Ji, and Neubig]{wang2025openhands}
Xingyao Wang, Boxuan Li, Yufan Song, Frank~F. Xu, Xiangru Tang, Mingchen Zhuge,
  Jiayi Pan, Yueqi Song, Bowen Li, Jaskirat Singh, Hoang~H. Tran, Fuqiang Li,
  Ren Ma, Mingzhang Zheng, Bill Qian, Yanjun Shao, Niklas Muennighoff, Yizhe
  Zhang, Binyuan Hui, Junyang Lin, Robert Brennan, Hao Peng, Heng Ji, and
  Graham Neubig.
\newblock Openhands: An open platform for {AI} software developers as
  generalist agents.
\newblock In \emph{The Thirteenth International Conference on Learning
  Representations}, 2025.
\newblock URL \url{https://openreview.net/forum?id=OJd3ayDDoF}.

\bibitem[Wei et~al.(2022)Wei, Wang, Schuurmans, Bosma, ichter, Xia, Chi, Le,
  and Zhou]{NEURIPS2022_9d560961}
Jason Wei, Xuezhi Wang, Dale Schuurmans, Maarten Bosma, brian ichter, Fei Xia,
  Ed~Chi, Quoc~V Le, and Denny Zhou.
\newblock Chain-of-thought prompting elicits reasoning in large language
  models.
\newblock In S.~Koyejo, S.~Mohamed, A.~Agarwal, D.~Belgrave, K.~Cho, and A.~Oh,
  editors, \emph{Advances in Neural Information Processing Systems}, volume~35,
  pages 24824--24837. Curran Associates, Inc., 2022.
\newblock URL
  \url{https://proceedings.neurips.cc/paper_files/paper/2022/file/9d5609613524ecf4f15af0f7b31abca4-Paper-Conference.pdf}.

\bibitem[Xia et~al.(2026)Xia, Chen, Wang, Liu, Zeng, Wang, Han, Zhou, Zhao,
  Chen, Zheng, Xie, and Yao]{xia2026skillrlevolvingagentsrecursive}
Peng Xia, Jianwen Chen, Hanyang Wang, Jiaqi Liu, Kaide Zeng, Yu~Wang, Siwei
  Han, Yiyang Zhou, Xujiang Zhao, Haifeng Chen, Zeyu Zheng, Cihang Xie, and
  Huaxiu Yao.
\newblock Skillrl: Evolving agents via recursive skill-augmented reinforcement
  learning, 2026.
\newblock URL \url{https://arxiv.org/abs/2602.08234}.

\bibitem[Yang et~al.(2025{\natexlab{a}})Yang, Li, Yang, Zhang, Hui, Zheng, Yu,
  Gao, Huang, Lv, Zheng, Liu, Zhou, Huang, Hu, Ge, Wei, Lin, Tang, Yang, Tu,
  Zhang, Yang, Yang, Zhou, Zhou, Lin, Dang, Bao, Yang, Yu, Deng, Li, Xue, Li,
  Zhang, Wang, Zhu, Men, Gao, Liu, Luo, Li, Tang, Yin, Ren, Wang, Zhang, Ren,
  Fan, Su, Zhang, Zhang, Wan, Liu, Wang, Cui, Zhang, Zhou, and
  Qiu]{yang2025qwen3technicalreport}
An~Yang, Anfeng Li, Baosong Yang, Beichen Zhang, Binyuan Hui, Bo~Zheng, Bowen
  Yu, Chang Gao, Chengen Huang, Chenxu Lv, Chujie Zheng, Dayiheng Liu, Fan
  Zhou, Fei Huang, Feng Hu, Hao Ge, Haoran Wei, Huan Lin, Jialong Tang, Jian
  Yang, Jianhong Tu, Jianwei Zhang, Jianxin Yang, Jiaxi Yang, Jing Zhou,
  Jingren Zhou, Junyang Lin, Kai Dang, Keqin Bao, Kexin Yang, Le~Yu, Lianghao
  Deng, Mei Li, Mingfeng Xue, Mingze Li, Pei Zhang, Peng Wang, Qin Zhu, Rui
  Men, Ruize Gao, Shixuan Liu, Shuang Luo, Tianhao Li, Tianyi Tang, Wenbiao
  Yin, Xingzhang Ren, Xinyu Wang, Xinyu Zhang, Xuancheng Ren, Yang Fan, Yang
  Su, Yichang Zhang, Yinger Zhang, Yu~Wan, Yuqiong Liu, Zekun Wang, Zeyu Cui,
  Zhenru Zhang, Zhipeng Zhou, and Zihan Qiu.
\newblock Qwen3 technical report, 2025{\natexlab{a}}.
\newblock URL \url{https://arxiv.org/abs/2505.09388}.

\bibitem[Yang et~al.(2024)Yang, Wang, Lu, Liu, Le, Zhou, and
  Chen]{ICLR2024_3339f19c}
Chengrun Yang, Xuezhi Wang, Yifeng Lu, Hanxiao Liu, Quoc~V Le, Denny Zhou, and
  Xinyun Chen.
\newblock Large language models as optimizers.
\newblock In B.~Kim, Y.~Yue, S.~Chaudhuri, K.~Fragkiadaki, M.~Khan, and Y.~Sun,
  editors, \emph{International Conference on Learning Representations}, volume
  2024, pages 12028--12068, 2024.
\newblock URL
  \url{https://proceedings.iclr.cc/paper_files/paper/2024/file/3339f19c5fcee3ad74502947a32be9e6-Paper-Conference.pdf}.

\bibitem[Yang et~al.(2025{\natexlab{b}})Yang, Lieret, Jimenez, Wettig,
  Khandpur, Zhang, Hui, Press, Schmidt, and
  Yang]{yang2025swesmithscalingdatasoftware}
John Yang, Kilian Lieret, Carlos~E. Jimenez, Alexander Wettig, Kabir Khandpur,
  Yanzhe Zhang, Binyuan Hui, Ofir Press, Ludwig Schmidt, and Diyi Yang.
\newblock Swe-smith: Scaling data for software engineering agents,
  2025{\natexlab{b}}.
\newblock URL \url{https://arxiv.org/abs/2504.21798}.

\bibitem[Yao et~al.(2023)Yao, Zhao, Yu, Du, Shafran, Narasimhan, and
  Cao]{yao2023react}
Shunyu Yao, Jeffrey Zhao, Dian Yu, Nan Du, Izhak Shafran, Karthik~R Narasimhan,
  and Yuan Cao.
\newblock React: Synergizing reasoning and acting in language models.
\newblock In \emph{The Eleventh International Conference on Learning
  Representations}, 2023.
\newblock URL \url{https://openreview.net/forum?id=WE_vluYUL-X}.

\bibitem[Zelikman et~al.(2022)Zelikman, Wu, Mu, and
  Goodman]{NEURIPS2022_639a9a17}
Eric Zelikman, Yuhuai Wu, Jesse Mu, and Noah Goodman.
\newblock Star: Bootstrapping reasoning with reasoning.
\newblock In S.~Koyejo, S.~Mohamed, A.~Agarwal, D.~Belgrave, K.~Cho, and A.~Oh,
  editors, \emph{Advances in Neural Information Processing Systems}, volume~35,
  pages 15476--15488. Curran Associates, Inc., 2022.
\newblock URL
  \url{https://proceedings.neurips.cc/paper_files/paper/2022/file/639a9a172c044fbb64175b5fad42e9a5-Paper-Conference.pdf}.

\bibitem[Zelikman et~al.(2024)Zelikman, Harik, Shao, Jayasiri, Haber, and
  Goodman]{zelikman2024quietstar}
Eric Zelikman, Georges~Raif Harik, Yijia Shao, Varuna Jayasiri, Nick Haber, and
  Noah Goodman.
\newblock Quiet-{ST}ar: Language models can teach themselves to think before
  speaking.
\newblock In \emph{First Conference on Language Modeling}, 2024.
\newblock URL \url{https://openreview.net/forum?id=oRXPiSOGH9}.

\bibitem[Zhao et~al.(2024)Zhao, Huang, Xu, Lin, Liu, and
  Huang]{Zhao_Huang_Xu_Lin_Liu_Huang_2024}
Andrew Zhao, Daniel Huang, Quentin Xu, Matthieu Lin, Yong-Jin Liu, and Gao
  Huang.
\newblock Expel: Llm agents are experiential learners.
\newblock \emph{Proceedings of the AAAI Conference on Artificial Intelligence},
  38\penalty0 (17):\penalty0 19632–19642, Mar. 2024.
\newblock \doi{10.1609/aaai.v38i17.29936}.
\newblock URL \url{https://ojs.aaai.org/index.php/AAAI/article/view/29936}.

\bibitem[Zhou et~al.(2026)Zhou, Zhang, Wang, Hao, Wang, Han, Yang, Wu, Pan,
  Fan, Tu, and Zhang]{zhou2026featurebench}
Qixing Zhou, JiaCheng Zhang, Haiyang Wang, Rui Hao, Jiahe Wang, Minghao Han,
  Yuxue Yang, Shuzhe Wu, Feiyang Pan, Lue Fan, Dandan Tu, and Zhaoxiang Zhang.
\newblock Featurebench: Benchmarking agentic coding for complex feature
  development.
\newblock In \emph{The Fourteenth International Conference on Learning
  Representations}, 2026.
\newblock URL \url{https://openreview.net/forum?id=41xrZ3uGuI}.

\bibitem[Zhou et~al.(2023)Zhou, Muresanu, Han, Paster, Pitis, Chan, and
  Ba]{zhou2023large}
Yongchao Zhou, Andrei~Ioan Muresanu, Ziwen Han, Keiran Paster, Silviu Pitis,
  Harris Chan, and Jimmy Ba.
\newblock Large language models are human-level prompt engineers.
\newblock In \emph{The Eleventh International Conference on Learning
  Representations}, 2023.
\newblock URL \url{https://openreview.net/forum?id=92gvk82DE-}.

\end{thebibliography}
\end{document}